\documentclass[10pt,twocolumn,letterpaper]{article}

\usepackage{cvpr}
\usepackage{times}
\usepackage{epsfig}
\usepackage{graphicx}
\usepackage{amsmath}
\usepackage{amssymb}
\usepackage{epstopdf}
\usepackage{amsthm}
\usepackage{multirow}
\usepackage[font=scriptsize]{subfig}
\usepackage{array}
\usepackage{threeparttable}
\usepackage{colortbl,booktabs}
\usepackage[linesnumbered,ruled]{algorithm2e}
\usepackage{url}
\usepackage{color}

\usepackage[breaklinks=true,bookmarks=false]{hyperref}

\DeclareMathOperator*{\argmax}{argmax}

\theoremstyle{definition}

\let\oldnl\nl
\newcommand{\nonl}{\renewcommand{\nl}{\let\nl\oldnl}}
\cvprfinalcopy 

\def\cvprPaperID{****} 

\begin{document}

\title{Jointly Deep Multi-View Learning for Clustering Analysis}


\author{Bingqian Lin$^1$\footnotemark[2]\quad\quad Yuan Xie$^2$\footnotemark[2]\quad\quad  Yanyun Qu$^1$\footnotemark[1]\quad\quad  Cuihua Li$^1$\quad\quad  Xiaodan Liang$^3$\\
$^1$Xiamen University\quad\quad $^2$Chinese Academy of Sciences\quad\quad $^3$Sun Yat-sen University
}

\maketitle
\renewcommand{\thefootnote}{\fnsymbol{footnote}} 
\footnotetext[2]{These authors contributed equally to this work.} 
\footnotetext[1]{Corresponding author is Yanyun Qu (yyqu@xmu.edu.cn).} 

\begin{abstract}
  In this paper, we propose a novel \textbf{J}oint framework for \textbf{D}eep \textbf{M}ulti-view  \textbf{C}lustering (\textbf{DMJC}), where multiple deep embedded features, multi-view fusion mechanism and clustering assignments can be learned simultaneously. Our key idea is that the joint learning strategy can sufficiently exploit clustering-friendly multi-view features and useful multi-view complementary information to improve the clustering performance. How to realize the multi-view fusion in such a joint framework is the primary challenge. To do so, we design two ingenious variants of deep multi-view joint clustering models under the proposed framework, where multi-view fusion is implemented by two different schemes. The first model, called DMJC-S, performs multi-view fusion in an implicit way via a novel multi-view soft assignment distribution. The second model, termed DMJC-T, defines a novel multi-view auxiliary target distribution to conduct the multi-view fusion explicitly. Both DMJC-S and DMJC-T are optimized under a KL divergence like clustering objective. Experiments on six challenging image datasets demonstrate the superiority of both DMJC-S and DMJC-T over single/multi-view baselines and the state-of-the-art multi-view clustering methods, which proves the effectiveness of the proposed DMJC framework. To our best knowledge, this is the first work to model the multi-view clustering in a deep joint framework, which will provide a meaningful thinking in unsupervised multi-view learning.

\end{abstract}

\section{Introduction}
As more and more real-world data are collected from diverse sources or obtained from different feature extractors, multi-view clustering has gained increasing attention in recent years. Its key idea is to exploit the complementary information among different views to boost the clustering performance. Early multi-view clustering methods utilize handcrafted features to perform clustering, {\it e.g.}, for image clustering, some approaches may use heterogeneous visual features such as SIFT \cite{SIFT}, LBP \cite{LBP}, and HOG \cite{HOG}.

With the development of deep learning, more powerful features can be learned by using various kinds of deep neural networks (DNN), such as stacked autoencoder (SAE) \cite{SAE}, variational autoencoder (VAE) \cite{VAE}, and convolutional autoencoder (CAE) \cite{DCEC}, which were proposed for unsupervised learning. DNN-based multi-view clustering methods \cite{DCCA, DCCAE, DGCCA,MVCDMF} have beat the traditional methods to a certain extent, by learning complex nonlinear transformations to obtain powerful multi-view features, and exploiting the effective relationship among multiple views such as canonical correlation \cite{DCCA, DCCAE, DGCCA}. However, either traditional multi-view methods or existing DNN-based ones perform clustering {\it in a separated way}, {\it i.e.}, multi-view features are firstly extracted, then traditional clustering such as K-means or spectral clustering is employed. This kind of separated learning strategy may bring the unsatisfactory clustering results since the relationship between the multi-view feature learning and clustering is not well exploited.

To solve above problems, we consider the following issue: {\it Can we develop a multi-view clustering approach, which can optimize the multi-view feature learning, multi-view fusion and clustering in a joint framework to improve the clustering performance?} We take inspiration from the recently proposed single-view joint clustering methods \cite{DEC, JULE, DCN}, which simultaneously optimize feature learning and clustering in a joint framework. These single-view based joint methods have shown great superiority over separated ones. However, they do not make use of the complementary information from multiple heterogeneous feature spaces, thus the clustering performance is still limited.

\begin{figure*}[htb]
\setlength{\abovecaptionskip}{4pt}
\setlength{\belowcaptionskip}{0pt}
\renewcommand{\figurename}{Figure}
\centering
\includegraphics[width=11cm,height=6.5cm]{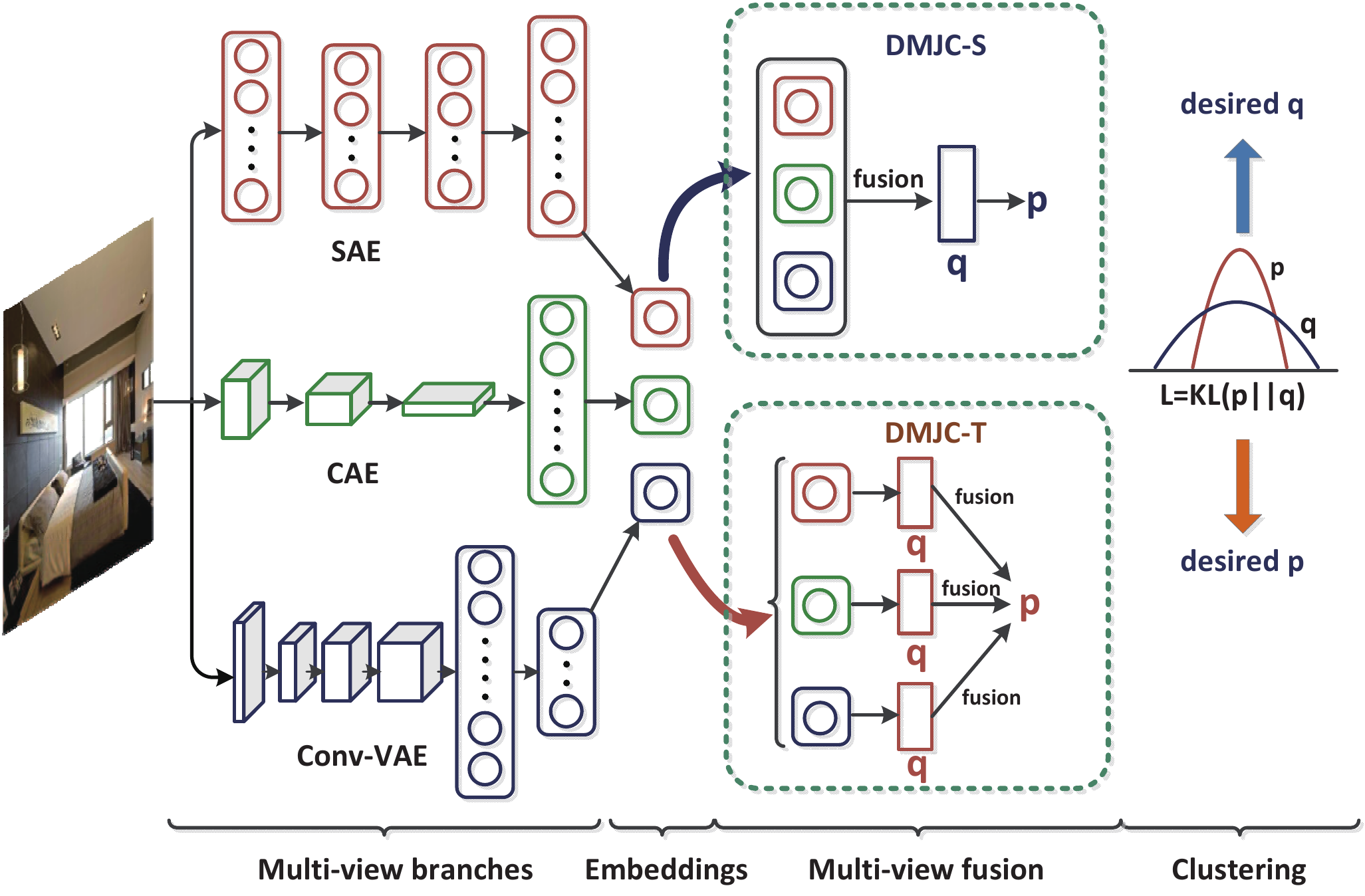}
\caption{The joint learning flowchart of the DMJC-S and DMJC-T. a) Multi-view branches: there are two techniques to construct multi-view branches, {\it i.e.}, feeding the same input into different network architectures or feeding distinct inputs into the same network architecture. b) Embeddings: the multi-view features for clustering are extracted from the embedding layers of multi-view branches. c) Multi-view fusion: for DMJC-S and DMJC-T, the multi-view fusion is conducted via the multi-view soft assignment distribution $\mathbf{q}$ and the multi-view auxiliary target distribution $\mathbf{p}$, respectively. d) Clustering: $\mathbf{q}$ or $\mathbf{p}$ is used to predict right clusters, {\it i.e.}, each sample is assigned to the cluster with the highest prediction probability.}
\label{fig:overview}
\end{figure*}

In this paper, we propose a novel deep multi-view joint clustering (DMJC) framework, which can optimize the multi-view feature representations, multi-view weighting mechanism, and image clustering simultaneously. In such a joint framework, effectively realizing the multi-view fusion to exploit the multi-view complementary information is vital and challenging. Considering the multi-view fusion in two different ways, we derive two variants of deep multi-view joint clustering models under the proposed framework, which are dubbed as DMJC-S and DMJC-T, respectively. DMJC-S realizes the implicit multi-view fusion scheme by defining a novel multi-view soft assignment distribution. DMJC-T, however, conducts the multi-view fusion explicitly through a novel multi-view auxiliary target distribution. Through the above two kinds of multi-view fusion schemes, multi-view complementary information can be effectively explored in both two models during the joint learning process. Fig. \ref{fig:overview} illustrates the joint learning flowchart of the proposed DMJC-S and DMJC-T. As shown in Fig. \ref{fig:overview}, the multi-view branches are based on various kinds of autoencoders.\footnote{Note that we only use the encoder parts of the autoencoders as the multi-view branches for joint learning.} Both DMJC-S and DMJC-T realize the joint learning through a KL divergence based clustering objective. To sum up, our main contributions are three-folds:

\begin{itemize}
\item We propose a novel joint framework for deep multi-view clustering (DMJC), which can learn multiple deep embedded features, multi-view fusion mechanism, and clustering assignments simultaneously. The proposed framework can effectively utilize the  multi-view complementary traits as well as clustering-friendly features to guide clustering more accurately.
\item Two variants of deep multi-view joint clustering models, termed DMJC-S and DMJC-T, are derived under the proposed framework. The DMJC-S and DMJC-T realize the multi-view fusion in an implicit and explicit way, respectively. Moreover, both the DMJC-S and DMJC-T can support end-to-end training.
\item Compared with the single/multi-view baselines and the state-of-the-art multi-view clustering methods, both the DMJC-S and DMJC-T can achieve the superior performance on challenging grayscale/three-channel benchmark image datasets, which demonstrates the effectiveness of the proposed DMJC framework.
\end{itemize}

\section{Related Work}

\paragraph{Multi-view clustering} Multi-view clustering can be roughly divided into discriminative approaches and generative approaches. The former attracts much more attentions, and can be further divided into three categories \cite{review-mvc}: (1) direct view combination; (2) common structure based methods; (3) view combination after projection. We mainly focus on the last two categories in this paper. One of the most representative methods for the second category is RMKMC \cite{RMKMC}, which defined a common indicator matrix across different views. The multi-view self-paced clustering (MSPL) \cite{MSPL}, which is an improvement of RMKMC, learned the multi-view model by taking the complexities of the samples and views into account. The most typical methods for the last category are canonical correlation analysis (CCA) based ones \cite{cca1,cca2,DCCA,DCCAE,DGCCA}, which project the multi-view high-dimensional data onto a low-dimensional subspace with CCA. The above mentioned multi-view clustering approaches can obtain powerful multi-view representations. However, they all employ a two-step pipeline to obtain the features and clusters, which often leads to the suboptimal clustering performance.

\paragraph{Deep joint clustering} Recently, clustering methods, which jointly learn feature representations and clustering via deep networks \cite{DEC,JULE,DCN}, have been proposed. Deep clustering network (DCN) \cite{DCN} proposed a joint dimensional reduction and K-means clustering framework, where the dimensional reduction model was based on deep neural network. Motivated by the t-SNE \cite{t-sne}, the deep embedded clustering (DEC) \cite{DEC} employed a deep stacked autoencoder (SAE) \cite{SAE} to initialize the feature extraction model, and then iteratively optimized a KL divergence based clustering objective with a self-training target distribution. These joint learning methods have shown great superiority beyond the separated ones. However, they only concentrate on the joint learning of single-view feature and clustering, and the joint learning of multi-view feature and clustering is under-explored. The proposed framework, different from aforementioned methods, learns multiple feature representations, multi-view fusion mechanism, and clustering assignments simultaneously, which can achieve superior clustering performance over single-view ones and acommodate well on large scale datasets.

\section{Deep Multi-View Joint Clustering}
Consider the problem of clustering a set of $N$ data points into $K$ clusters by using multi-view features $\{\mathbf{x}_i^{(v)} \in \mathbf{X}^{(v)}\}_{i=1}^{N}$, where the $\mathbf{X}^{(v)}$ denotes the original $v$-th feature space, $v=1,\ldots, V$. Each cluster is represented by a centroid $\boldsymbol{\mu}_j^{(v)}, j=1,\ldots, K$. Instead of conducting multi-view clustering directly in the \emph{original feature spaces} $\{\mathbf{X}^{(v)}\}_{i=1}^{N}$, we aim to transform the original features with non-linear mapping $f_{\boldsymbol{\theta}^{(v)}}^{(v)}: \mathbf{X}^{(v)} \rightarrow \mathbf{Z}^{(v)}$, where $\boldsymbol{\theta}^{(v)}$ represents the learnable hyperparameters of mapping functions for the $v$-th view, and $\mathbf{Z}^{(v)}$ is the latent \emph{embedded feature space}. Specifically, the non-linear mapping here is parameterized by the deep neural network.

The proposed framework simultaneously learns view-specific $K$ cluster centers $\boldsymbol{\mu}_j^{(v)}$ in the embedded feature space $\mathbf{Z}^{(v)}$, the parameters $\boldsymbol{\theta}^{(v)}$ of the deep network, and the multi-view fusion parameters in a unified way. Since the proposed framework is based on DEC, in the following, we will review DEC first and then elaborate the proposed DMJC-S and DMJC-T.

\subsection{Deep Embedded Clustering}

\begin{figure}[htb]
\centering
\includegraphics[width=0.45\textwidth]{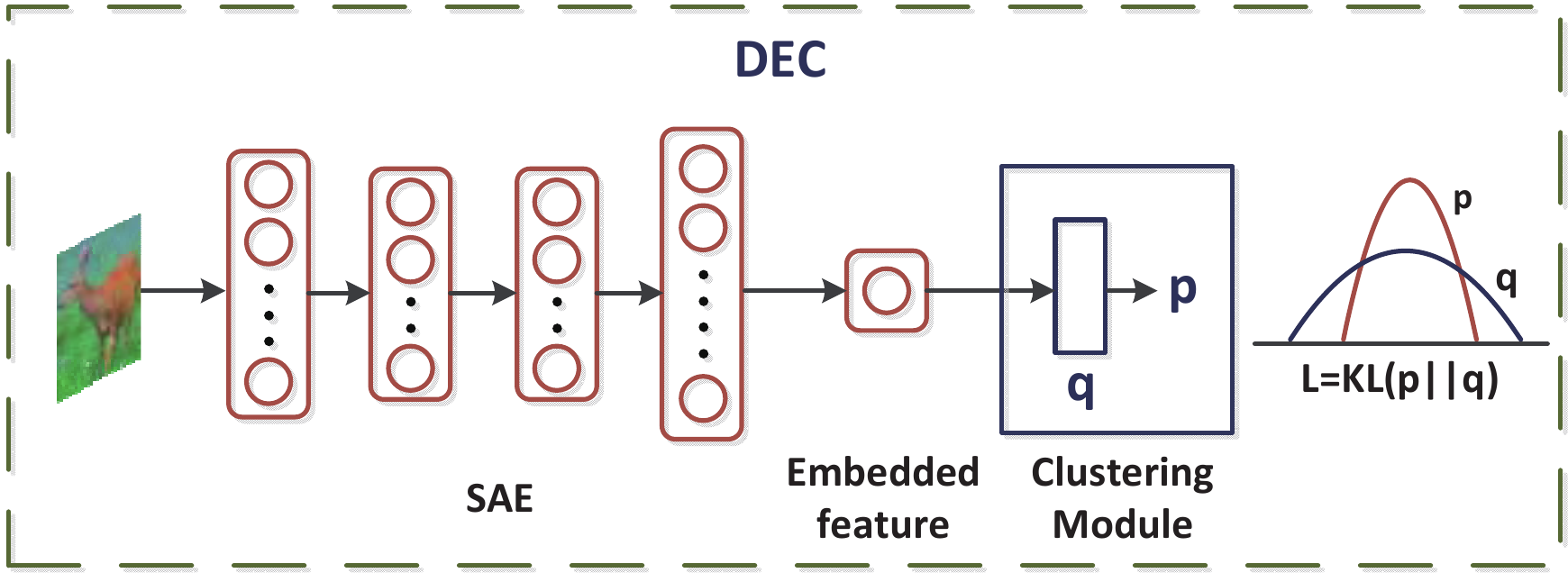}
\caption{The framework of DEC.}\label{DEC}
\end{figure}

The framework of DEC \cite{DEC} is shown in Fig. \ref{DEC}. As a single-view joint clustering method, DEC contains the parameter initialization stage and the joint learning stage based on SAE. After pretraining of the SAE, the initial estimation of network parameter $f_{\boldsymbol{\theta}}$ is obtained and the initial cluster centroids $\boldsymbol{\mu}_j$ is derived by performing K-means on the embedded feature $\mathbf{z}_i$. Then DEC adopts the Student's $t$-distribution as the soft assignment to measure the similarity between the embedded feature $\mathbf{z}_i$ and the cluster center $\boldsymbol{\mu}_j$:
\begin{equation}\label{view-specific-q}
    \mathbf{q}_{ij} = \frac{(1 + \|\mathbf{z}_i- \boldsymbol{\mu}_j\|^{2}/\alpha)^{-\frac{\alpha+1}{2}}}{\sum_{j'} (1 + \|\mathbf{z}_i - \boldsymbol{\mu}_{j'}\|^{2}/\alpha)^{-\frac{\alpha+1}{2}}},
\end{equation}
where $\alpha$ represents the degree of freedom of the Student's $t$-distribution. $\mathbf{q}_{ij}$ is referred to as the \emph{soft assignment} because it can be interpreted as the probability of assigning sample $i$ to the $j$-th cluster.
Since the label is unreachable, DEC iteratively refines the clusters with the help of a powerful auxiliary target distribution:
\begin{equation}\label{view-specific-p}
\begin{aligned}
    &\mathbf{p}_{ij} = \left(\mathbf{q}_{ij}\right)^{\gamma}, \gamma > 1, \text{s.t.} \quad \sum_{j}\mathbf{p}_{ij} = 1.
\end{aligned}
\end{equation}
This target distribution can guide the clustering by boosting the high confidence scores of the soft assignment. Based on the above soft assignment and auxiliary target, DEC optimizes a KL divergence based clustering loss:
\begin{equation}\label{obj-DEC}
\begin{aligned}
    \boldsymbol{\mathcal{L}}(\boldsymbol{\mu}_j, \boldsymbol{\theta}) = \min_{\boldsymbol{\mu}_j, \boldsymbol{\theta}} & \sum_{i}\sum_{j} \mathbf{p}_{ij} \log\frac{\mathbf{p}_{ij}}{\mathbf{q}_{ij}},
\end{aligned}
\end{equation}

Through above clustering objective, the embedded features $\mathbf{z}$ and clustering assignments $\boldsymbol{\mu}$ can be optimized jointly by using stochastic graident descent (SGD) with momentum. Note that the optimization of $\mathbf{z}$ is achieved by optimizing the network parameter $\boldsymbol{\theta}$.

Similar to DEC, the proposed framework also has parameters initialization stage and the optimization stage. Specifically, we use different autoencoders as different deep branches (views). After pretraining of multiple deep branches, we concatenate the embedded features $\mathbf{z}_i^{(v)}$ for the $v$-th view directly and perform K-means on the concatenated features to obtain initialized clustering centroids $\boldsymbol{\mu}_j^{(v)}$. By introducing a multi-view soft assignment distribution and a multi-view auxiliary target distribution respectively for DMJC-S and DMJC-T, the joint optimization process can be realized.

\subsection{DMJC-S: Implicit Multi-View Fusion in Soft Assignment Distribution}
\subsubsection{Problem Formulation}
In DMJC-S, a multi-view soft assignment distribution is defined as:
\begin{equation}\label{DMVC-MM-q}
    \mathbf{q}_{ij} = \frac{\sum_{v} \boldsymbol{\pi}_{j}^{(v)}(1 + \|\mathbf{z}_i^{(v)} - \boldsymbol{\mu}_j^{(v)}\|^{2}/\alpha)^{-\frac{\alpha+1}{2}}}{\sum_{j'}\sum_{v'} \boldsymbol{\pi}_{j'}^{(v')}(1 + \|\mathbf{z}_i^{(v')} - \boldsymbol{\mu}_{j'}^{(v')}\|^{2}/\alpha)^{-\frac{\alpha+1}{2}}},
\end{equation}
where the $\boldsymbol{\pi}_{j}^{(v)}$ denotes the importance weight that measures the importance of cluster center $j$ in $v$-th view for final clustering. This multi-view soft assignment distribution realizes the multi-view fusion by imposing the {\it implicit} multi-view constraint on the view-specific soft assignment, which is more powerful than single-view one in DEC. Since direct optimization of the objective function w.r.t the parameter $\boldsymbol{\pi}_{j}^{(v)}$ is difficult, we further represent the importance weight in terms of unconstrained weight $\mathbf{w}_{j}^{(v)}$ as:
\begin{equation}\label{softmax-weights}
    \boldsymbol{\pi}_{j}^{(v)} = \frac{e^{\mathbf{w}_{j}^{(v)}}}{\sum_{v'}e^{\mathbf{w}_{j}^{(v')}}}.
\end{equation}
The above equation restricts the importance weight $\boldsymbol{\pi}_{j}^{(v)}$ to be positive and $\sum_{v=1}^{V} \boldsymbol{\pi}_{j}^{(v)} = 1$. The unconstrained weight $\mathbf{w}_{j}^{(v)}$ can be easily learned by gradient descent. The derivation of auxiliary target distribution $\mathbf{p}_{ij}$ is the same as DEC. As a result, the DMJC-S model can be defined as:
\begin{equation}\label{obj-DMVC-MM}
\begin{aligned}
    \boldsymbol{\mathcal{L}}(\boldsymbol{\mu}_j^{(v)}, \boldsymbol{\theta}^{(v)}, \boldsymbol{W}) = \min_{\boldsymbol{\mu}_j^{(v)}, \boldsymbol{\theta}^{(v)}, \boldsymbol{W}} & \sum_{i}\sum_{j} \mathbf{p}_{ij} \log\frac{\mathbf{p}_{ij}}{\mathbf{q}_{ij}},
\end{aligned}
\end{equation}
where the $\boldsymbol{W}$ denotes a matrix containing $K\times V$ unconstrained weights $\mathbf{w}_j^{(v)}$.

\subsubsection{Optimization Procedure}
The initialized importance weight $\boldsymbol{\pi}_{j}^{(v)}$ is set as $1/V$ for the $j$ clustering centroid in the $v$-th view. After parameters initialization, we jointly optimize the cluster centers $\boldsymbol{\mu}_j^{(v)}$, hyperparameter $\boldsymbol{\theta}^{(v)}$ of multiple deep networks, and the importance weight matrix $\boldsymbol{W}$ by using SGD with momentum. The gradients of $\boldsymbol{\mathcal{L}}$ w.r.t embedding feature $\mathbf{z}_i^{(v)}$, each cluster center $\boldsymbol{\mu}_j^{(v)}$, and the element of $\boldsymbol{W}$, {\it i.e.,} $\mathbf{w}_{j}^{(v)}$ can be calculated as:
\begin{equation}\label{}
    \frac{\partial\boldsymbol{\mathcal{L}}}{\partial\mathbf{z}_i^{(v)}} = \frac{2}{\alpha}\sum_{j} \frac{\partial\boldsymbol{\mathcal{L}}}{\partial \mathbf{d}_{ij}^{(v)}} (\mathbf{z}_i^{(v)} - \boldsymbol{\mu}_j^{(v)}),
\end{equation}

\begin{equation}\label{}
    \frac{\partial\boldsymbol{\mathcal{L}}}{\partial\boldsymbol{\mu}_j^{(v)}} = -\frac{2}{\alpha}\sum_{i} \frac{\partial\boldsymbol{\mathcal{L}}}{\partial \mathbf{d}_{ij}^{(v)}} (\mathbf{z}_i^{(v)} - \boldsymbol{\mu}_j^{(v)}),
\end{equation}

\begin{equation}\label{}
    \frac{\partial\boldsymbol{\mathcal{L}}}{\partial\mathbf{w}_{j}^{(v)}} = \boldsymbol{\pi}_{j}^{(v)}\left[ \frac{\partial \boldsymbol{\mathcal{L}}}{\partial \boldsymbol{\pi}_{j}^{(v)}} - \sum_{v'} \boldsymbol{\pi}_{j}^{(v')} \frac{\partial \boldsymbol{\mathcal{L}}}{\partial\boldsymbol{\pi}_{j}^{(v')}} \right],
\end{equation}
where $\mathbf{d}_{ij}^{(v)}=\| \mathbf{z}_{i}^{(v)} - \boldsymbol{\mu}_{j}^{(v)} \|^{2}/\alpha$. The gradients $\partial\boldsymbol{\mathcal{L}}/\partial\mathbf{z}_i^{(v)}$ will be passed down to the deep network to further compute the hyperparameter $\partial\boldsymbol{\mathcal{L}}/\partial\boldsymbol{\theta}^{(v)}$ via the standard backpropagation algorithm. The detailed gradient derivations are given in the supplemental material.

Once all the parameters have been optimized, the predicted label for the unlabeled data will be derived by:
\begin{equation}\label{decision-DMVC-MM}
    \mathbf{y}_i = \argmax_{j} \mathbf{q}_{ij}.
\end{equation}

\subsection{DMJC-T: Explicit Multi-View Fusion in Auxiliary Target Distribution}
\subsubsection{Problem Formulation}
In DMJC-T, we first derive the view-specific soft assignment distribution $\mathbf{q}_{ij}^{(v)}$ and auxiliary target distribution $\mathbf{p}_{ij}^{(v)}$ like DEC, then a multi-view auxiliary target distribution is calculated by:
\begin{equation}\label{mvtd}
    \mathbf{p}_{ij} = \sum_{v} \mathbf{w}_v \mathbf{p}_{ij}^{(v)},
\end{equation}
where $\mathbf{w}_v$ is the multi-view weight for the $v$-th view, with the strong constraints that $\mathbf{w}_v > 0$ and $\sum_{v} \mathbf{w}_v = 1$. This multi-view auxiliary target distribution  realizes the multi-view fusion by imposing the {\it explicit} multi-view constraint on the view-specific auxiliary target distribution, which can provide more powerful guidance than single-view one in DEC. With this new multi-view auxiliary target, the DMJC-T model is defined as:
\begin{equation}\label{obj-DMVC-ME}
\begin{aligned}
    &\boldsymbol{\mathcal{L}}(\boldsymbol{\mu}_j^{(v)}, \boldsymbol{\theta}^{(v)}, \mathbf{w}) = \\
    &\min_{\boldsymbol{\mu}_j^{(v)}, \boldsymbol{\theta}^{(v)}, \mathbf{w}}  \sum_{v'}\sum_{i}\sum_{j}  \left\{ (\sum_{v} \mathbf{w}_v \mathbf{p}_{ij}^{(v)}) \log\frac{(\sum_{v} \mathbf{w}_v \mathbf{p}_{ij}^{(v)})}{\mathbf{q}_{ij}^{(v')}}\right\}
    \\ &+ \lambda \|\mathbf{w}\|_{2}^{2},\\
    & \text{s.t.} \quad \mathbf{w}_v > 0, \quad \sum_{v} \mathbf{w}_v = 1.
\end{aligned}
\end{equation}
Note that the second term, {\it i.e.,} $l_2$-norm of $\mathbf{w}$, is introduced as a regularization term to avoid the trivial solution (only one of $\mathbf{w}_v$ is equal to $1$), which means that only one view works, while the contributions of the other views vanish. The parameter $\lambda$ is used to balance the effects of the two parts in (\ref{obj-DMVC-ME}).

\subsubsection{Optimization Procedure}
The initialized multi-view weight for the $v$-th view is set as $1/V$. After initialization, we conduct the joint optimization over network parameters $\boldsymbol{\theta}^{(v)}$, view-specific cluster centers $\boldsymbol{\mu}_j^{(v)}$, and multi-view weight $\mathbf{w}$. We adopt alternative optimization to solve this problem. Concretely, in every iteration, we fix $\mathbf{w}$ to update $\boldsymbol{\mu}_i^{(v)}$ and $\boldsymbol{\theta}^{(v)}$. After that, we fix $\boldsymbol{\mu}_i^{(v)}$ and $\boldsymbol{\theta}^{(v)}$ to optimize $\mathbf{w}$. The details of optimization procedure are given below:\\

\textbf{Optimize $\boldsymbol{\mu}_i^{(v)}$ and $\boldsymbol{\theta}^{(v)}$ by fixing $\mathbf{w}$:} since $\mathbf{w}$ is fixed, for the $v$-th view, the objective function $\boldsymbol{\mathcal{L}}$ is reduced to a KL divergence. The cluster center $\boldsymbol{\mu}_i^{(v)}$ and hyperparameter
$\boldsymbol{\theta}^{(v)}$ can be jointly optimized by using SGD with momentum. Similar to DEC,
the gradients of $\boldsymbol{\mathcal{L}}$ w.r.t embedded feature $\mathbf{z}_i^{(v)}$ and each cluster center $\boldsymbol{\mu}^{(v)}$ can be calculated as:
\begin{equation}\label{derivative1}
\begin{aligned}
     \frac{\partial\boldsymbol{\mathcal{L}}}{\partial\mathbf{z}_i^{(v)}} =
    & \frac{\alpha + 1}{\alpha} \sum_{j} \left(1 + \frac{\|\mathbf{z}_i^{(v)} - \boldsymbol{\mu}_j^{(v)}\|^2}{\alpha}\right)^{-1} \\
    & (\mathbf{p}_{ij} - \mathbf{q}_{ij}^{(v)})(\mathbf{z}_i^{(v)} - \boldsymbol{\mu}_j^{(v)}),
\end{aligned}
\end{equation}
\begin{equation}\label{derivative2}
\begin{aligned}
     \frac{\partial\boldsymbol{\mathcal{L}}}{\partial\boldsymbol{\mu}_j^{(v)}} =
    & -\frac{\alpha + 1}{\alpha} \sum_{i} \left(1 + \frac{\|\mathbf{z}_i^{(v)} - \boldsymbol{\mu}_j^{(v)}\|^2}{\alpha}\right)^{-1} \\
    & (\mathbf{p}_{ij} - \mathbf{q}_{ij}^{(v)})(\mathbf{z}_i^{(v)} - \boldsymbol{\mu}_j^{(v)}).
\end{aligned}
\end{equation}

\textbf{Optimize $\mathbf{w}$ by fixing $\boldsymbol{\mu}_i^{(v)}$ and $\boldsymbol{\theta}^{(v)}$:} since $\boldsymbol{\mu}_i^{(v)}$ and $\boldsymbol{\theta}^{(v)}$ are fixed, $\mathbf{p}_{ij}^{(v)}$ and $\mathbf{q}_{ij}^{(v)}$ are also fixed. The objective function w.r.t $\mathbf{w}$ is formulated as:

\begin{equation}\label{obj-DMVC-ME-w}
\begin{aligned}
    &\boldsymbol{\mathcal{L}}(\mathbf{w}) = \min_{\mathbf{w}}  \sum_{v'}\sum_{i}\sum_{j} \mathbf{p}_{ij} \log\frac{\mathbf{p}_{ij}}{\mathbf{q}_{ij}^{(v')}}
    + \lambda \|\mathbf{w}\|_{2}^{2},\\
    & \text{s.t.} \quad \mathbf{w}_v > 0, \quad \sum_{v} \mathbf{w}_v = 1.
\end{aligned}
\end{equation}

It can be easily proved that $\boldsymbol{\mathcal{L}}(\mathbf{w})$ is convex w.r.t $\mathbf{w}$, since the KL divergence is convex w.r.t $\mathbf{p}_{ij}$, which is linear w.r.t $\mathbf{w}$. The $l_2$-norm is also convex w.r.t $\mathbf{w}$. This subproblem can be efficiently solved by accelerated proximal gradient (APG) method \cite{APG1}.

Similarly to DMJC-S, the predicted label for the unlabeled data is derived by:
\begin{equation}\label{decision-DMVC-ME}
    \mathbf{y}_i = \argmax_{j} \mathbf{p}_{ij}.
\end{equation}

\begin{figure}[htb]
\centering
\includegraphics[width=0.45\textwidth]{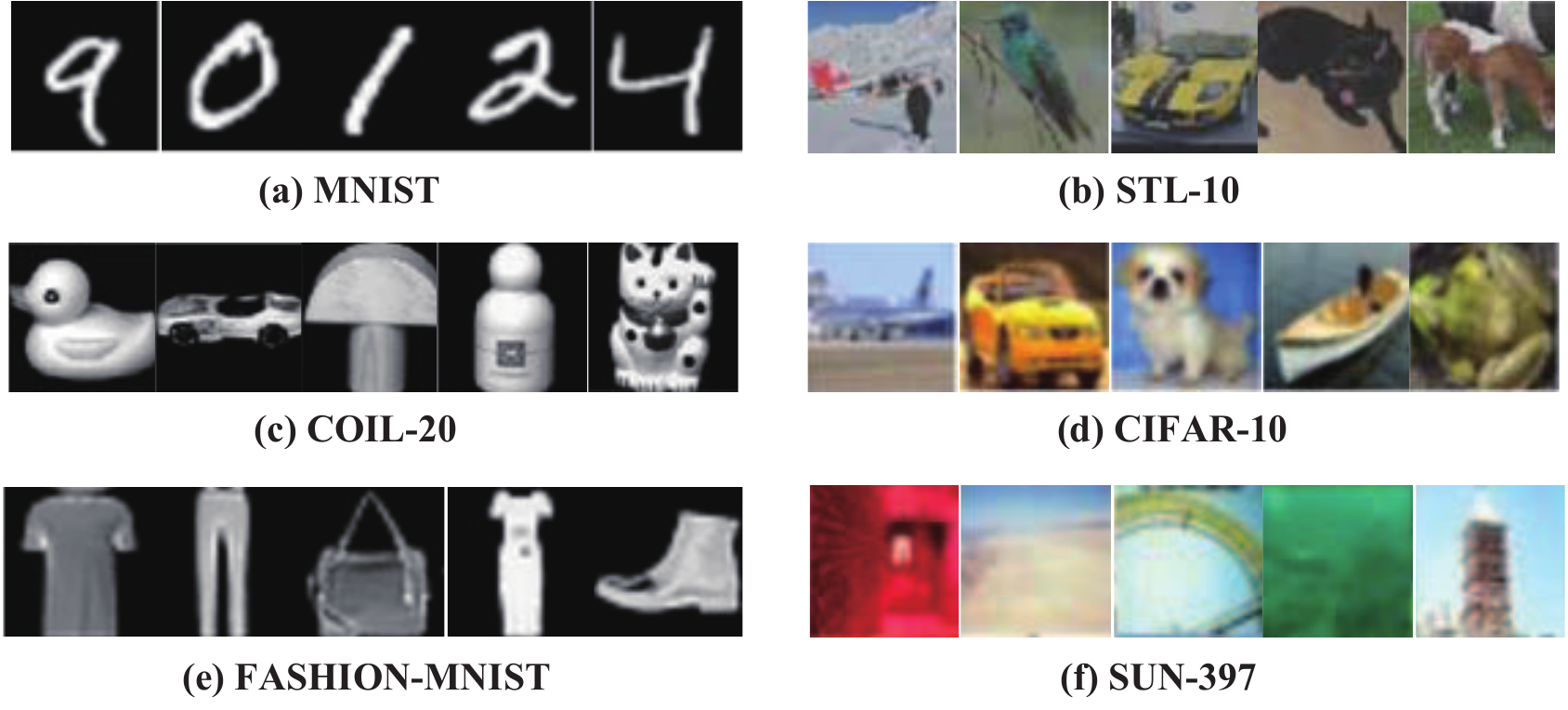}
\caption{The examples of six image datasets. (a),(c),(e) are gray-scale image datasets, (b),(d),(f) are three-channel image datasets.}\label{datasets_examples}
\end{figure}

\subsection{Relationship between DMJC-S and DMJC-T}
In DMJC-S, by formulating the multi-view soft assignment distribution as the weighted sum of view-specific soft assignment, the multi-view fusion is realized in an implicit way, {\it which associates the view-specific clustering centroid in all views.} Alternatively, in DMJC-T, the multi-view fusion is explicitly conducted on the auxiliary target distribution, meaning that the soft assignment in each view is the same as that in DEC, {\it leading to the independent relationship of the view-specific centroid in all views.} For performance comparison, the DMJC-T can usually achieve better clustering results than the DMJC-S (see Table \ref{experiment-results1}), which shows that the explicit fusion scheme is more powerful than the implicit counterpart. However, since DMJC-S benefits from a simpler optimization strategy, its training process is more efficient. While the optimization problem (\ref{obj-DMVC-ME-w}) for the DMJC-T is still a bottleneck for large-scale dataset.

\section{Experiments}
\subsection{Experimental Settings}
\textbf{Datasets.} To demonstrate the performance of the proposed framework, we evaluate the DMJC-S and DMJC-T on six popular image datasets: MNIST \cite{mnist}, fashion-MNIST \cite{fashion-mnist}, COIL-20 \cite{DBC}, STL-10 \cite{stl10}, CIFAR-10 \cite{cifar10} and SUN-397 \cite{sun397}. \textbf{MNIST} consists of $70K$ grayscale hand-written digit images with a size of $28\times28$ pixels from 10 categories. \textbf{FASHION-MNIST} contains $70K$ fashion product images from $10$ classes, with the same image size to MNIST. \textbf{COIL-20} collects $1440$ $128\times128$ grayscale object images of $20$ categories viewed from varying angles. \textbf{STL-10} incorporates $13K$ three-channel images with the size of $96\times96$ pixels from $10$ object classes. \textbf{CIFAR-10} is similar to STL-10, which comprises $60K$ three-channel images of $10$ object categories, and the image size is $32\times32$ pixels. \textbf{SUN-397} dataset includes $108,754$ scene images from $397$ categories. The number of images in SUN-397 varies across categories, with at least $100$ images per category. The statistics and examples of six image datasets are shown in Table \ref{dataset} and Fig. \ref{datasets_examples}. Note that the training and testing images in each dataset are jointly utilized for clustering.

\renewcommand{\arraystretch}{1.5}
\setlength{\abovecaptionskip}{0pt}%
\setlength{\belowcaptionskip}{0pt}%
\begin{table}[!ht]
  \centering
  \fontsize{7}{7}\selectfont
\caption{The summary of data statistics.}
\label{dataset}
\begin{center}
\begin{tabular}{|c|c|c|c|}
\hline
Datasets&Numbers&Classes&Image sizes\\ \hline
MNIST&70000&10&(28,28,1)\\
FASHION-MNIST&70000&10&(28,28,1)\\
COIL-20&1440&20&(128,128,1)\\
STL-10&13000&10&(96,96,3)\\
CIFAR-10&60000&10&(32,32,3)\\
SUN-397&108754&397&\verb|\| \\
\hline
\end{tabular}
\end{center}
\end{table}

\textbf{Network configurations.} We employ the stacked autoencoder (SAE) \cite{SAE}, convolutional variational autoencoder (Conv-VAE) \cite{Conv-VAE1}, and convolutional autoencoder (CAE) \cite{DCEC} as three single-view deep network branches for gray-scaled image datasets MNIST, FASHION-MNIST and COIL-20. For Conv-VAE and CAE, the original image is taken as the network input. Since the layers of SAE are densely connected, the raw image is converted into a long vector to feed into the input layer.

For three-channeled image datasets {\it i.e.,} STL-10, CIFAR10 and SUN-397, we use two SAE networks with different input features and one variational autoencoder (VAE) network as the single-view deep network branches. To pursue better performance, we employ deep features as the input of different branches. Specifically, the deep features are extracted from the fully-connected layers of four powerful networks, {\it i.e.,} the AlexNet \cite{AlexNet}, VGG-VD \cite{VGG-VD}, ResNet50 \cite{ResNet}, and Inception-V3 \cite{Inception-V3}, all of which are pre-trained on ILSVRC2012 \cite{imagenet}. A detailed configurations of the single-view network branches in our models, the corresponding encoder architectures (the decoder parts are symmetric to the encoder parts) and the network inputs on different datasets are given in Table \ref{implementation-detail}. Note that the selections of network or network input are not important, since we mainly focus on exploiting the multi-view complementary information to improve the clustering performance.

\textbf{Evaluation metrics.} To measure the clustering performance of different algorithms, we adopt three standard clustering criteria, {\it i.e.,} clustering accuracy (ACC), Normalized Mutual Information (NMI) and Adjusted Rand Index (ARI). These scores range in $[0,1]$, and for each of the metrics, the higher it is, the better the performance is.

\begin{table}[!htbp]
\fontsize{6}{6}\selectfont
\setlength{\abovecaptionskip}{0pt}%
\setlength{\belowcaptionskip}{0pt}%
\caption{The detailed configurations of the proposed models on different datasets. ({\it The configurations of DMJC-S and DMJC-T are kept the same})}\label{implementation-detail}
\begin{center}
\begin{tabular}{|c|p{0.08\textwidth}|p{0.12\textwidth}|p{0.1\textwidth}|}
\hline
\centering{\textsf{Datasets}}& \centering{\textsf{Branches}}& \centering{\textsf{Encoder Architecture}}& \centering{\textsf{Network input}} \tabularnewline   \hline
\multirow{3}{*}[-20pt]{\shortstack{MNIST\\FASHION-MNIST\\COIL-20}}&SAE(View1)&500-500-2000-10&vectorized raw image\\ \cline{2-4}
&Conv-VAE(View2)&conv1(2$\times$2$\times$1)-conv2(2$\times$2$\times$6)-conv3(3$\times$3$\times$20)-
conv4(3$\times$3$\times$60)-flatten-256-10&raw image pixels\\ \cline{2-4}
&CAE(View3)&conv1(5$\times$5$\times$32, strides=2)-conv2(5$\times$5$\times$64, strides=2)-
conv3(3$\times$3$\times$128, strides=2)-flatten-10&raw image pixels\\ \hline
\multirow{3}{*}{\shortstack{STL-10\\CIFAR-10}}&SAE(View1)&500-500-2000-10&VGG16 feature\\ \cline{2-4}
&SAE(View2)&500-500-2000-10&ResNet50 feature\\ \cline{2-4}
&VAE(View3)&500-256-50&Inception-V3 feature\\ \hline
\multirow{3}{*}{SUN397}&SAE(View1)&2000-500-50-10&Alexnet feature \\ \cline{2-4}
&SAE(View2)&2000-500-50-10&Inception-V3 feature\\ \cline{2-4}
&VAE(View3)&2000-256-10&VGG19 feature\\ \hline
\end{tabular}
\end{center}
\end{table}

\renewcommand{\arraystretch}{1.5}
\begin{table*}[!htb]
\fontsize{5}{5}\selectfont
\caption{The quantative results on six image datasets. The best two scores are shown in \textcolor{red}{red} and \textcolor{blue}{blue} colors, respectively.}
\label{experiment-results1}
\begin{center}
\begin{tabular}{|c|c|c|c|c|c|c|c|c|c|c|c|c|c|c|c|c|c|c|}
\hline
\multirow{2}{*}{Method}&\multicolumn{3}{c|}{MNIST \cite{mnist}}&\multicolumn{3}{c|}{FASHION-MNIST \cite{fashion-mnist}}&\multicolumn{3}{c|}{COIL-20 \cite{DBC}}&\multicolumn{3}{c|}{STL-10 \cite{stl10}}&\multicolumn{3}{c|}{CIFAR-10 \cite{cifar10}}&\multicolumn{3}{c|}{SUN-397 \cite{sun397}}\cr\cline{2-19}
&ACC&NMI&ARI&ACC&NMI&ARI&ACC&NMI&ARI&ACC&NMI&ARI&ACC&NMI&ARI&ACC&NMI&ARI\cr
    \hline
    \hline
RMKMC \cite{RMKMC}&0.8255&0.7995&0.7524&0.5912&0.6169&0.4636&0.5799&0.7487&0.5275&0.8344&0.8273&0.7635&0.5714&0.4688&0.3679&0.2684&0.5049&0.1679\\
MSPL \cite{MSPL}&0.8717&0.8147&0.7892&0.5607&0.6068&0.4457&0.5992&0.7623&0.5608&0.8108&0.8220&0.7402&0.7156&0.5948&0.5174&0.2704&0.5072&0.1643\\
DCCA \cite{DCCA}&0.3155&0.2086&0.1272&0.4105&0.4028&0.2342&0.5512&0.7013&0.4600&0.8411&0.7477&0.6917&0.4242&0.3385&0.2181&0.2275&0.4749&0.1257\\
DCCAE \cite{DCCAE}&0.3029&0.2038&0.1274&0.4109&0.3836&0.2303&0.5551&0.7058&0.4667&0.8235&0.7273&0.6632&0.3960&0.3226&0.2034&0.2259&0.4731&0.1236\\
DGCCA \cite{DGCCA}&0.4714&0.3840&0.2789&0.4765&0.4827&0.3105&0.5337&0.6762&0.4370&0.8960&0.8218&0.7970&0.4703&0.3577&0.2634&0.1422&0.3578&0.0635\\
S-View-1&0.8901&0.8229&0.8034&0.5855&0.6240&0.4610&0.6957&0.7875&0.6200&0.8169&0.7415&0.6617&0.2983&0.1772&0.1024&0.1759&0.4137&0.0842\\
S-View-2&0.7971&0.6979&0.6545&0.5443&0.5044&0.3721&0.6555&0.7717&0.5922&0.8327&0.7562&0.6914&0.4331&0.3359&0.2243&0.2742&0.5272&0.1760\\
S-View-3&0.8888&0.8068&0.7888&0.5793&0.6343&0.4632&0.636&0.7607&0.5650&0.8224&0.7305&0.6411&0.7162&0.5714&0.5013&0.2205&0.4628&0.1186\\
S-All-Views&0.9062&0.8286&0.8181&0.5993&0.6348&0.4756&0.6737&0.7845&0.6124&0.9298&0.8674&0.8525&0.7552&0.6069&0.5428&0.2572&0.5108&0.1540\\
J-View-1&0.9378&0.9123&0.9018&0.5982&0.6407&0.4820&0.6870&0.7973&0.6261&0.8372&0.7570&0.7038&0.3035&0.1832&0.1094&0.1837&0.4116&0.0884\\
J-View-2&0.8929&0.8644&0.8236&0.5886&0.6011&0.4434&0.6675&0.7899&0.6112&0.8466&0.7658&0.7145&0.4423&0.3447&0.2398&0.2751&0.5198&\textcolor{blue}{0.1734}\\
J-View-3&0.9071&0.8251&0.8208&0.5462&0.5841&0.4136&0.5597&0.7383&0.4969&0.8939&0.8212&0.7863&0.7613&\textcolor{blue}{0.6424}&0.5721&0.2193&0.4645&0.1174\\
DMJC-S&\textcolor{blue}{0.9579}&\textcolor{blue}{0.9269}&\textcolor{blue}{0.9263}&\textcolor{red}{0.6208}&\textcolor{red}{0.6475}&\textcolor{red}{0.4958}&\textcolor{red}{0.7148}&\textcolor{blue}{0.8033}&\textcolor{blue}{0.6434}&\textcolor{blue}{0.9374}&\textcolor{blue}{0.8765}&\textcolor{blue}{0.8663}&\textcolor{blue}{0.7633}&0.6401&\textcolor{blue}{0.5776}&\textcolor{blue}{0.2795}&\textcolor{blue}{0.5313}&0.1753\\
DMJC-T&\textcolor{red}{0.9603}&\textcolor{red}{0.9312}&\textcolor{red}{0.9316}&\textcolor{blue}{0.6087}&\textcolor{blue}{0.6442}&\textcolor{blue}{0.4864}&\textcolor{blue}{0.7122}&\textcolor{red}{0.8080}&\textcolor{red}{0.6524}&\textcolor{red}{0.9471}&\textcolor{red}{0.8919}&\textcolor{red}{0.8872}&\textcolor{red}{0.7954}&\textcolor{red}{0.6784}&\textcolor{red}{0.6291}&\textcolor{red}{0.2884}&\textcolor{red}{0.5736}&\textcolor{red}{0.1812}\\
\hline
\end{tabular}
\end{center}
\end{table*}

\textbf{Compared methods.} We compare the proposed DMJC-S and DMJC-T with one multi-view baseline, the related single-view baselines, and several state-of-the-art multi-view clustering approaches. For the single-view competitors, we consider their learning strategies in two different ways. Similar to DEC \cite{DEC}, the first strategy (named as \textbf{J-View-i}) {\it jointly} learns feature representations and clustering, and $i$ denotes the $i$-th branch in our models. The other strategy (denoted as \textbf{S-View-i}) performs clustering {\it separately}, {\it i.e.}, it first trains the single deep autoencoder, then performs K-means based on the embedded feature. As for the multi-view baseline (named as \textbf{S-All-Views}), we train all the deep branches in our model independently, then concatenate all the embedded features directly to perform K-means.

Moreover, several representative multi-view clustering algorithms are also compared with our models, including DCCA \cite{DCCA}, DCCAE \cite{DCCAE}, DGCCA \cite{DGCCA}, RMKMC \cite{RMKMC} and MSPL \cite{MSPL}. The first three methods belong to DNN-based multi-view approaches, and the last two are chosen since they can be run on large-scale datasets.

\textbf{Implementation details.} The experiments are implemented on a workstation with Intel (R) Core (TM) i7-7700K @ 4.20GHz CPU, $120$GB RAM, and GeForce GTX 1080 GPU (8GB caches). For DMJC-S and DMJC-T, we apply the Adam optimizer \cite{Adam} in the pretraining process for all the deep views and the Adagrad optimizer \cite{AdaGrad} in the joint optimization process. The batch size for pretraining and joint optimization is set to $256$, and the maximum numbers of training epochs in the optimization process are set to $100$ and $2000$ for grayscale and three-channel image datasets, respectively. The parameter $\alpha$ and $\gamma$ in the calculation of $\mathbf{q}_{ij}$ and $\mathbf{p}_{ij}$ are set as $1$ and $2$. For DMJC-T, the balance factor $\lambda$ is set as $2.0\times 10^4$ for all datasets.

For the fairness of comparison, we set the same configurations, including optimizers, initializers, and batch sizes for the proposed models and the \textbf{J-View-i} algorithms on each dataset. Moreover, the pretrained network parameters of \textbf{J-View-i} are kept the same as our models. We also use the pretrained network parameters in our models to extract clustering features for \textbf{S-View-i}, \textbf{S-All-Views}, RMKMC and MSPL.

Since DCCA and DCCAE can only deal with two views, we choose the best two views in our models according to their performance as the two branches for DCCA and DCCAE. After multi-view feature learning, we concatenate the embedding features in two branches to perform K-means. For DGCCA, we use the shared representations to perform K-means directly. The pretrained network parameters of DCCA, DCCAE and DGCCA are also kept consistent with our models.

\subsection{Performance Comparison}
In this section, we report the quantitative results of different clustering methods. For each method, we run the experiments for 10 times on each dataset and get the average results. The results on six image datasets are shown in Table \ref{experiment-results1}, where we can find that the proposed DMJC-S and DMJC-T significantly outperform other clustering algorithms with all the metrics on both gray-scale and three-channel image datasets. When compared with  single-view baselines \textbf{J-View-i}, we can observe that in most cases, the clustering performance of the proposed two multi-view models will not be affected by the worst view and what's more, they can achieve higher accuracy than the best view. Moreover, we can also find that the joint learning strategy is generally superior to the separated ones (see \textbf{J-View-i} vs. \textbf{S-View-i}, DMJC-S/DMJC-T vs. \textbf{S-All-Views}), which shows the effectiveness of joint learning.

When the number of classes increases (see the results on COIL-20 and SUN-397), the proposed DMJC-S and DMJC-T can still keep the superior performance over other methods. In particular, from the experimental results on the challenging scene dataset SUN-397 (containing \textbf{100K+} samples with \textbf{397} categories), we can find that the proposed two models also show promising clustering ability when compared with single/multi-view baselines.

\subsection{Ablation Study}
\begin{figure}
\setlength{\abovecaptionskip}{3pt}
\setlength{\belowcaptionskip}{3pt}
  \centering
  \renewcommand{\figurename}{Figure}
  \vspace{-0.01in}
  \subfloat[ACC increments on STL10]{
    \includegraphics[width=3.6cm,height=2.6cm]{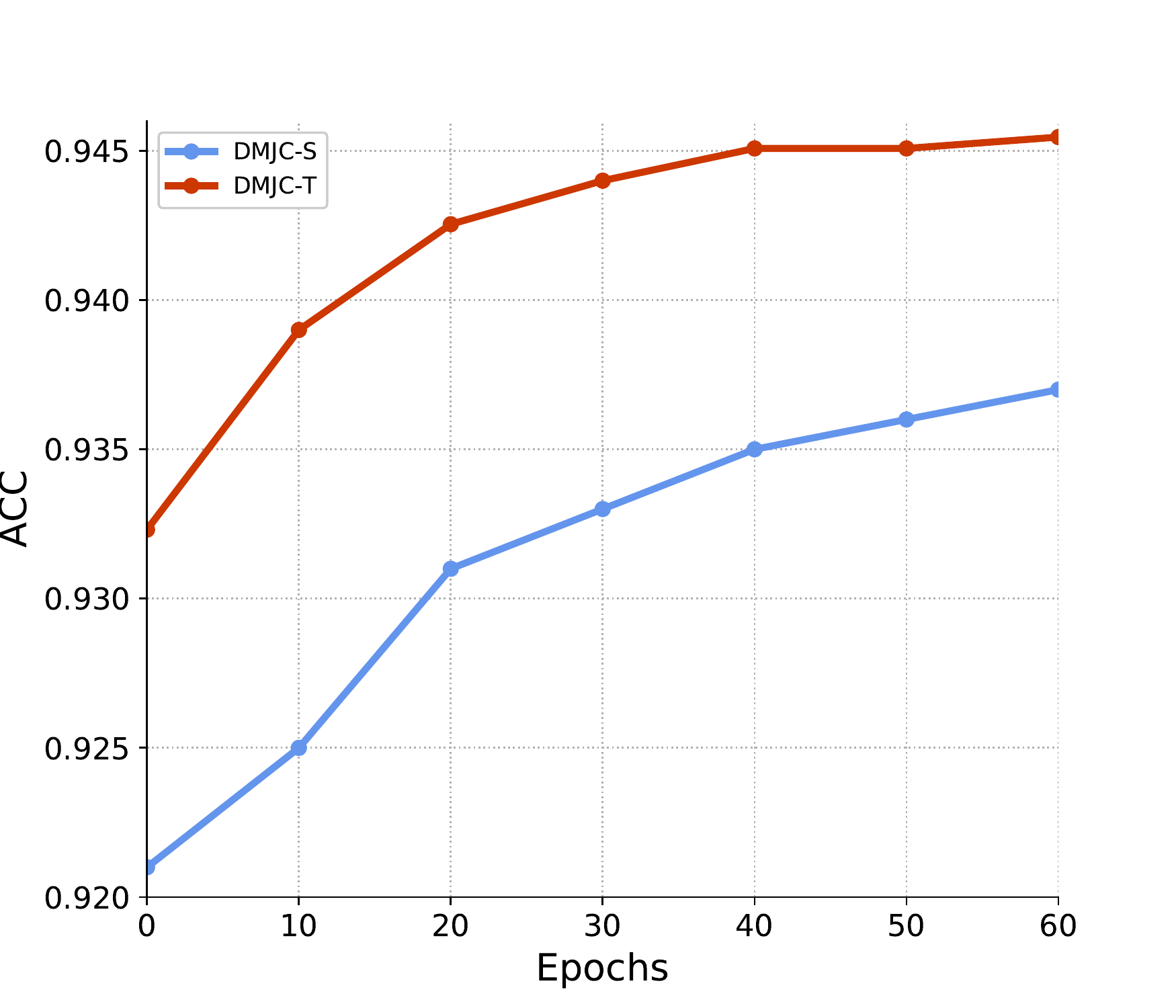}}
  \hspace{-0.01in}
  \subfloat[ACC increments on STL10]{
    \includegraphics[width=3.6cm,height=2.6cm]{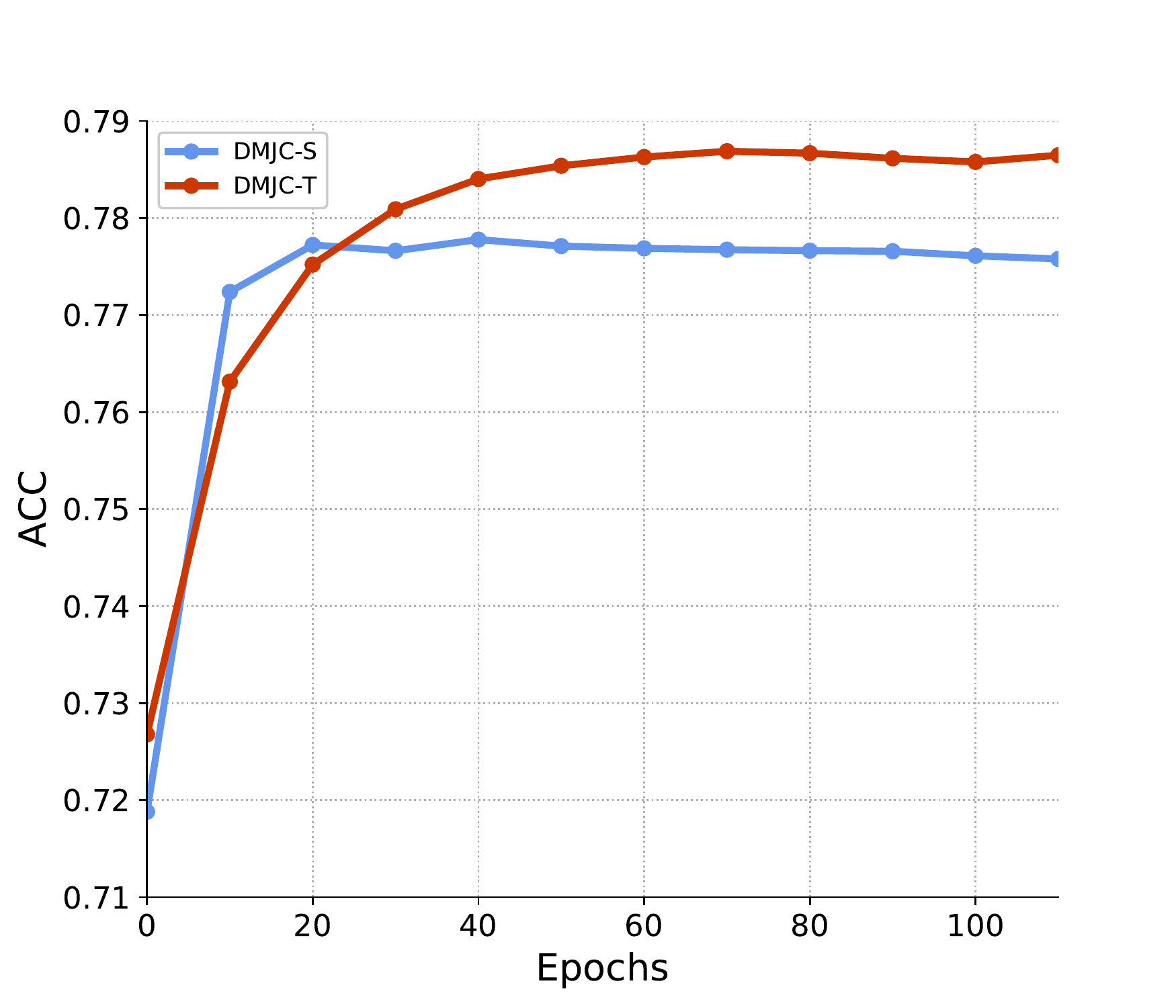}}\\
  \caption{The ACC increments during the optimization process of the DMJC-S and DMJC-T on STL10 and CIFAR10.}
  \label{accuracy-boost1} 
\end{figure}

\textbf{Contributions of multi-view branches.} Since the proposed models are based on joint framework, we investigate the effect of multi-view joint learning and the impact of different views on our models. To do the former, we report the change of the ACC in the  optimization process. As illustrated in Fig. \ref{accuracy-boost1}, the ACC results of both DMJC-S and DMJC-T are boosted during the optimization process on both STL-10 and CIFAR-10. Note that similar observations can be obtained on other datasets. The apparent performance gain proves that the proposed DMJC framework can effectively capture the relationship among large scale unlabeled data.

\begin{figure*}
\setlength{\abovecaptionskip}{3pt}
\setlength{\belowcaptionskip}{3pt}
  \centering
  \renewcommand{\figurename}{Figure}
  \vspace{-0.01in}
  \subfloat[View-specific ACC on STL10]{
    \includegraphics[width=4.5cm,height=2.5cm]{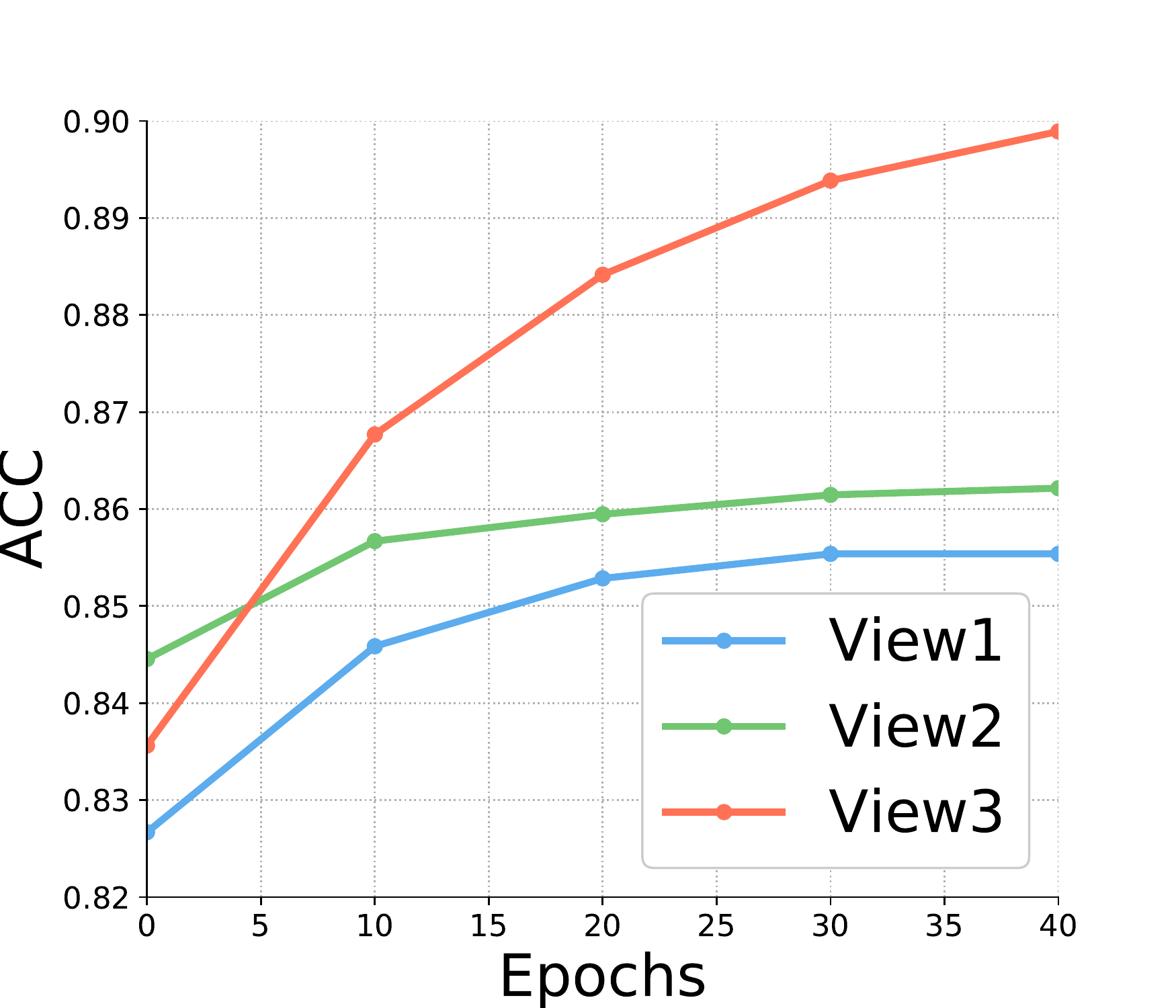}}
  \hspace{-0.01in}
  \subfloat[Weight variation on STL10 (DMJC-S)]{
    \includegraphics[width=4.5cm,height=2.5cm]{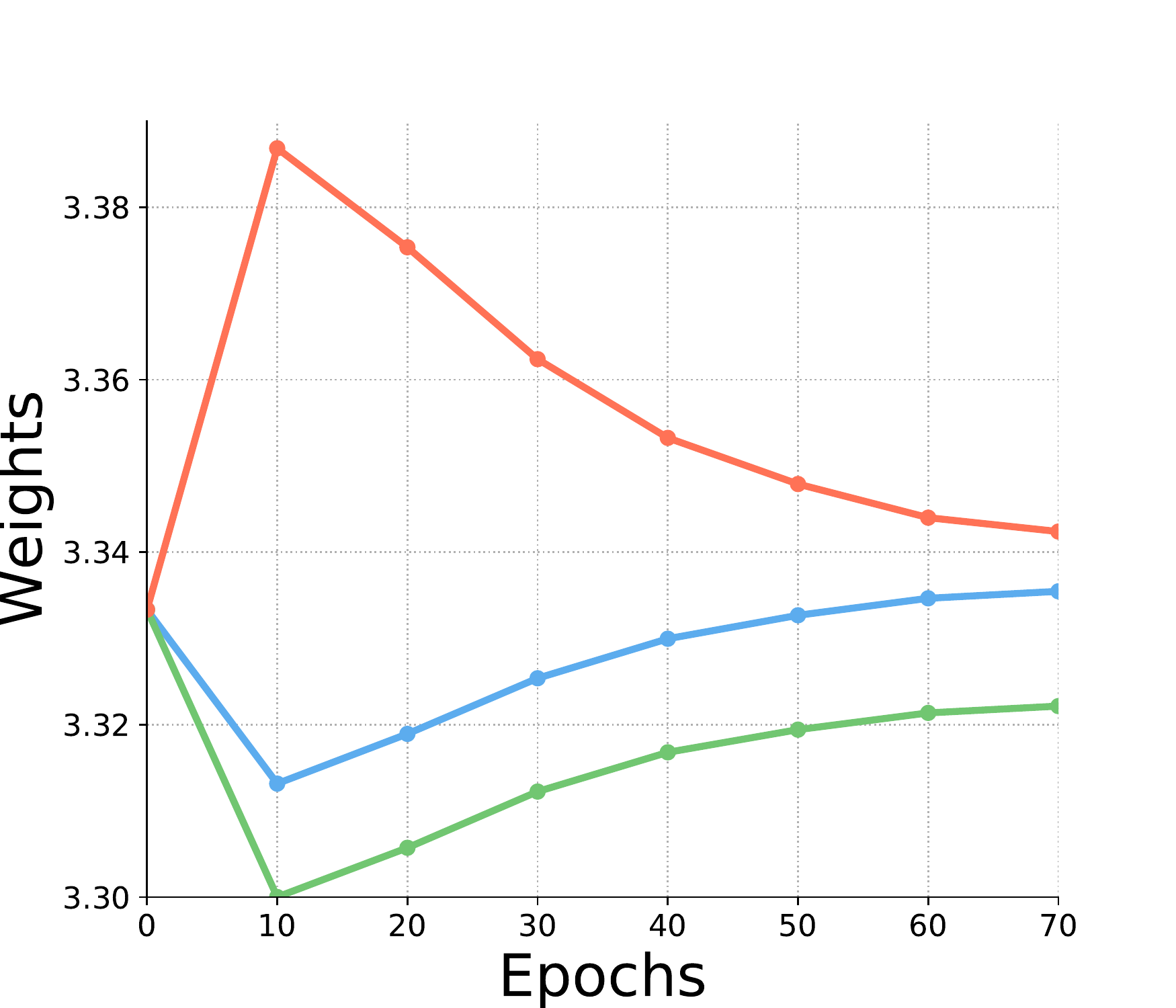}}
  \hspace{-0.01in}
  \subfloat[Weight variation on STL10 (DMJC-T)]{
    \includegraphics[width=4.5cm,height=2.5cm]{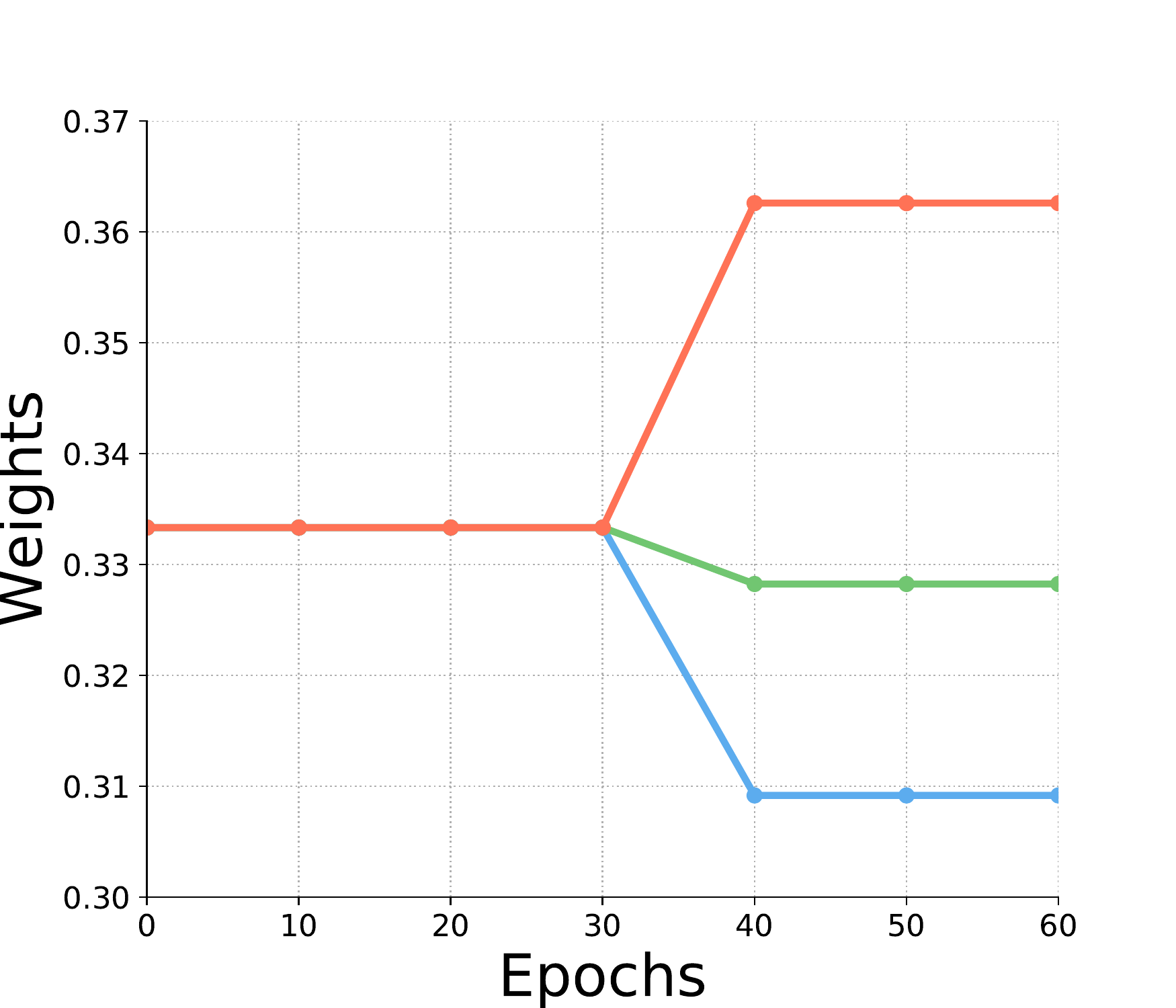}}\\
  \vspace{-0.15in}
  \subfloat[View-specific ACC on CIFAR10]{
    \includegraphics[width=4.5cm,height=2.5cm]{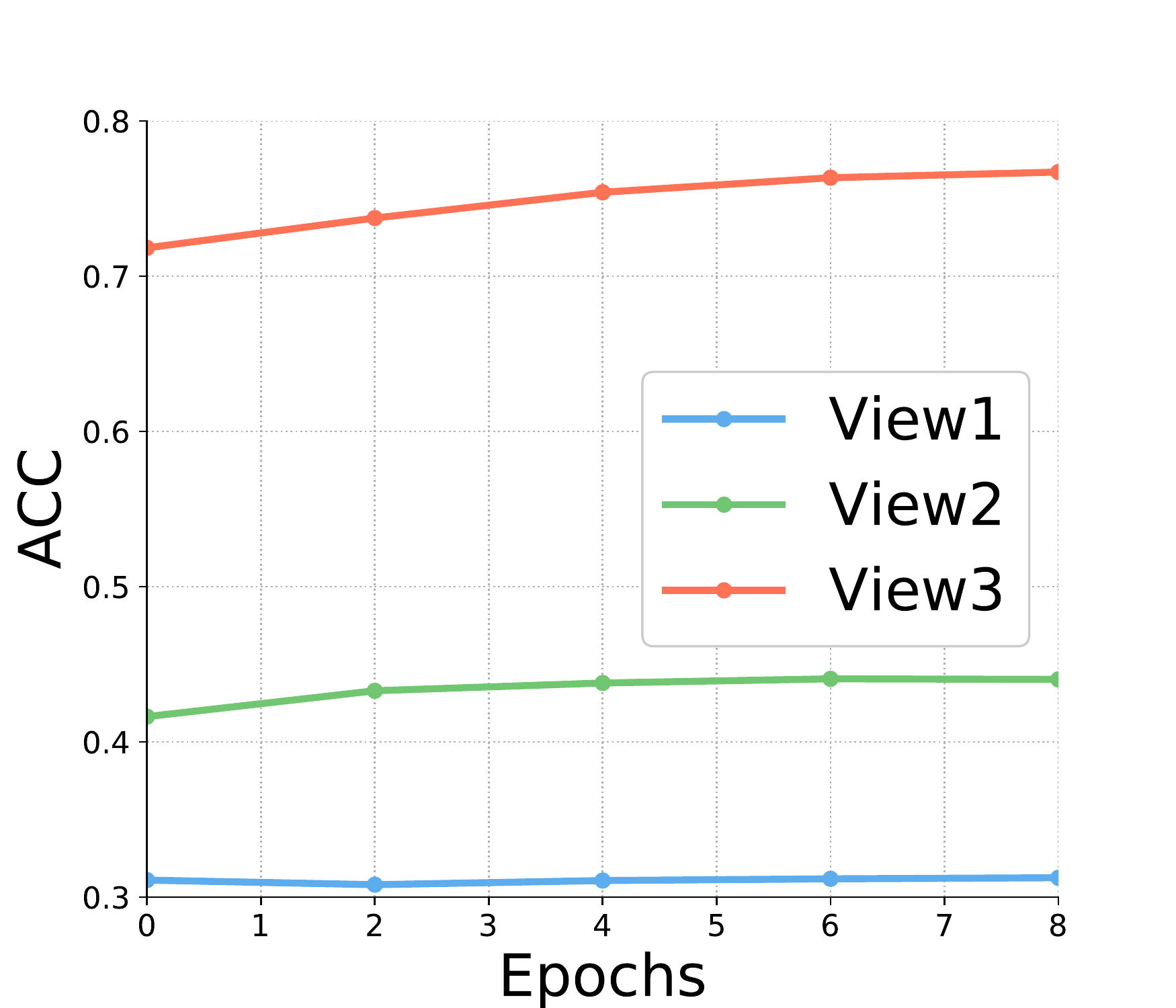}}
  \hspace{-0.01in}
  \subfloat[Weight variation on CIFAR10 (DMJC-S)]{
    \includegraphics[width=4.5cm,height=2.5cm]{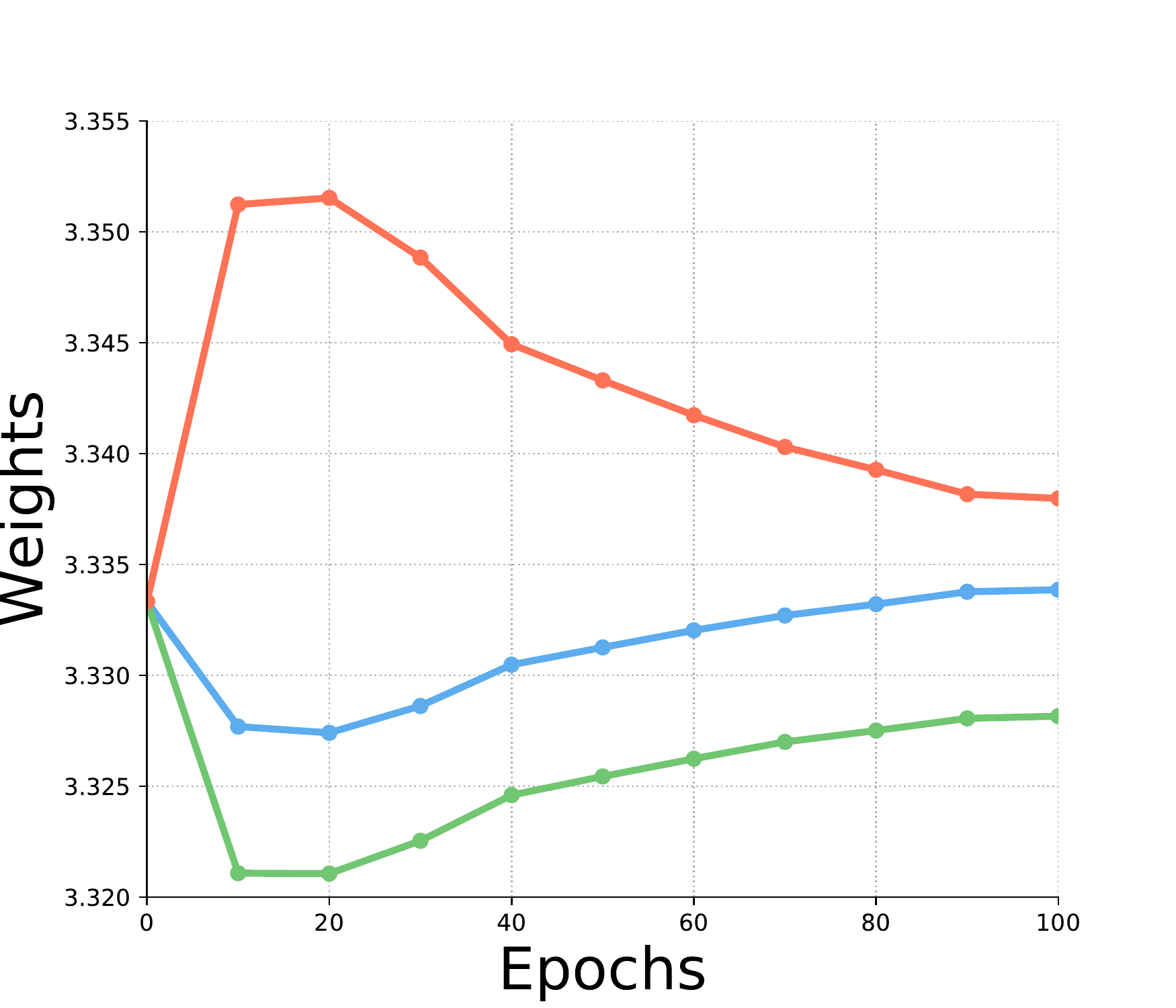}}
  \hspace{-0.01in}
  \subfloat[Weight variation on CIFAR10 (DMJC-T)]{
    \includegraphics[width=4.5cm,height=2.5cm]{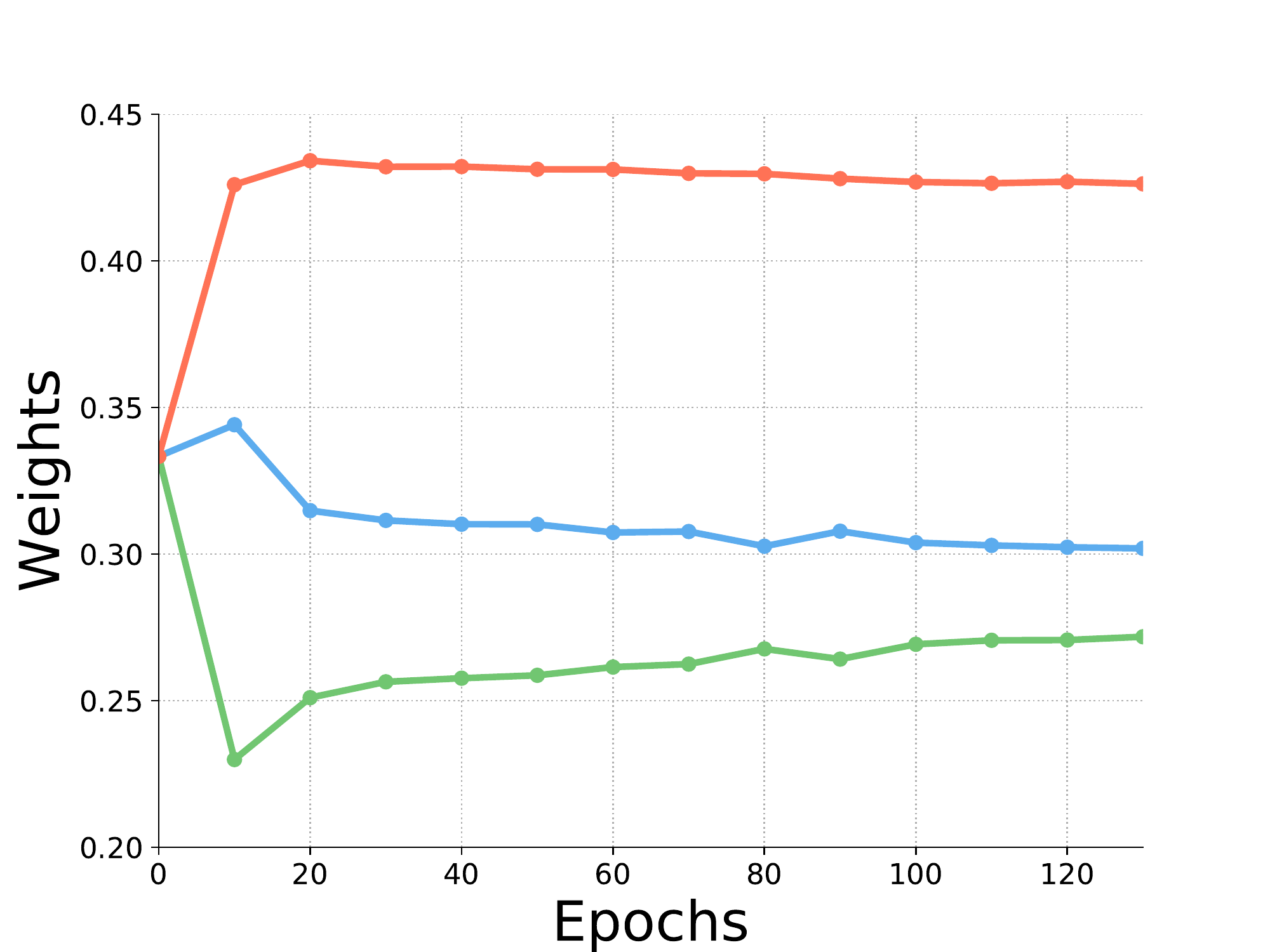}}
  \caption{The accuracy increments and weight variations of all the views for the DMJC-S and DMJC-T on STL-10 and CIFAR-10.}\label{accuracy-boost2} 
\end{figure*}

To do the latter, we show the variation of the multi-view weights, as well as the change of the ACC results of different views in our models during the optimization process on STL-10 and CIFAR-10. Note that the multi-view weight shown here for DMJC-S is the sum value for specific view. The related results are presented in Fig. \ref{accuracy-boost2}, where we can obviously find the view-specific boost from all the branches, which supports the overall performance gains of DMJC-S and DMJC-T. Furthermore, the order of color curves indicates that, the best view (the red curve in Fig. \ref{accuracy-boost2} (a) and (d)) will dominate the final result in the learning process (the red curve in Fig. \ref{accuracy-boost2} (b), (c), (e) and (f)), while other views' impacts can not be ranked according to their individual performance. Probably this is determined by the importance degree of complementary information in specific view.

\textbf{Sensitivity for initialization.} As said before, the joint learning framework contains the initialization stage and the optimization stage. Thus, we investigate the sensitivity for initialization to validate whether the proposed models can achieve stable performance in most cases. Therefore, we report the ACC variances for the proposed models and \textbf{J-View-i} on one grayscale image dataset (FASHION-MNIST) and two three-channel image datasets (STL-10 and CIFAR-10) in Table \ref{robustness-for-initialization}. We run all the methods for $50$ times to obtain the variance on each dataset. From Table \ref{robustness-for-initialization}, we can find that relatively low variances are kept by DMJC-S and DMJC-T, indicating the good stability of the proposed models.

\renewcommand{\arraystretch}{1.5}
\begin{table}[!htb]
\fontsize{6}{6}\selectfont
\caption{The ACC variances of DMJC-S and DMJC-T on FASHION-MNIST, STL-10 and CIFAR-10.}\label{robustness-for-initialization}
\begin{center}
\begin{tabular}{l|ccc}
\hline
\toprule[1pt]
Method&FASHION-MNIST&STL10&CIFAR10\\ \hline
J-View-1(CAE)&0.001753&0.001977&0.000225\\
J-View-2(Conv-VAE)&0.000689&0.001672&0.000163\\
J-View-3(SAE)&0.00162&0.000241&0.001481\\ \hline
DMJC-S&0.000976&1.06e-5&3.65e-5\\
DMJC-T&0.000929&1.20e-6&4.66e-5\\
\hline
\bottomrule[1pt]
\end{tabular}
\end{center}
\end{table}

\textbf{Robustness for integrating different views.} In the aforementioned experiments, we fix the number of views (usually three) in the proposed models. To study the impact of the number of views on our models, we construct the proposed models with two, three, and four network branches, where the configurations are (Conv-VAE + SAE), (CAE + Conv-VAE + SAE), and (CAE + Conv-VAE + SAE + VAE), respectively. For the limitation of space, we only report the related clustering results of our models and the corresponding \textbf{J-View-i} on MNIST, which are shown in Table \ref{robustness-for-number-of-views}.

Compared with the J-View-2 and J-View-3, the two-view DMJC-S and DMJC-T achieve better performances ($93.84\%$ and $93.88\%$ vs $89.29\%$ and $90.71\%$ for ACC), which indicates that two types of features generated by the Conv-VAE and SAE can complement each other well under the proposed framework. Moreover, compared with three-view counterparts ($93.84\%$ and $93.88\%$ vs $95.79\%$ and $96.03\%$ for ACC), the two-view models can still achieve highly competitive results even though a good branch, {\it i.e.,} CAE, has been removed. Furthermore, when adding a {\it degenerative view} (VAE, the ACC of J-View-4 is $70.59\%$) into our three-view models, the effects on clustering performance are limited with a reasonable margin ($88.63\%$ and $88.28\%$ for ACC). These experimental results sufficiently demonstrate the robustness of the proposed DMJC-S and DMJC-T for incorporating different views.

\renewcommand{\arraystretch}{1.5}
\begin{table}[!htb]
\fontsize{6}{6}\selectfont
\caption{The clustering results of DMJC-S and DMJC-T with incorporating different types of views.}\label{robustness-for-number-of-views}
\begin{center}
\begin{tabular}{l|ccc}
\hline
\toprule[1pt]
Methods&ACC&NMI&ARI\\ \hline
J-View-1 (CAE)&0.9378&0.9123&0.9018\\
J-View-2 (Conv-VAE)&0.8929&0.8644&0.8236\\
J-View-3 (SAE)&0.9071&0.8251&0.8208\\
J-View-4 (VAE)&0.7059&0.6930&0.5769\\ \hline
DMJC-S (Conv-VAE + SAE)&0.9384&0.9109&0.8995\\
DMJC-T (Conv-VAE + SAE)&0.9388&0.9090&0.8993\\ \hline
DMJC-S (CAE + Conv-VAE + SAE)&0.9579&0.9269&0.9263\\
DMJC-T (CAE + Conv-VAE + SAE)&0.9603&0.9312&0.9316\\ \hline
DMJC-S (CAE + Conv-VAE + SAE + VAE)&0.8863&0.8777&0.8438 \\
DMJC-T (CAE + Conv-VAE + SAE + VAE)&0.8828&0.8702&0.8370 \\
\hline
\bottomrule[1pt]
\end{tabular}
\end{center}
\end{table}

\section{Conclusions}
In this paper, we propose a novel deep multi-view joint clustering framework, which learns multiple deep embedded features, multi-view weighting mechanism and clustering assignments concurrently. Two elegant deep multi-view joint clustering models are derived under the proposed framework, where implicit and explicit multi-view fusion schemes are realized, respectively. Experimental results on both grayscale/three-channel image datasets fully substantiate the superiority of the proposed two models, which shows the effectiveness of the proposed joint learning framework.

{\small
\bibliographystyle{ieee}
\bibliography{egbib}
}

\end{document}